# DNN-BASED SEMANTIC MODEL FOR RESCORING N-BEST SPEECH RECOGNITION LIST

*Dominique Fohr, Irina Illina*

Université de Lorraine, CNRS, Inria, Loria, F-54000 Nancy, France
{dominique.fohr,irina.illina}@loria.fr

## ABSTRACT

The word error rate (WER) of an automatic speech recognition (ASR) system increases when a mismatch occurs between the training and the testing conditions due to the noise, etc. In this case, the acoustic information can be less reliable. This work aims to improve ASR by modeling long-term semantic relations to compensate for distorted acoustic features. We propose to perform this through rescoring of the ASR N-best hypotheses list. To achieve this, we train a deep neural network (DNN). Our DNN rescoring model is aimed at selecting hypotheses that have better semantic consistency and therefore lower WER. We investigate two types of representations as part of input features to our DNN model: static word embeddings (from *word2vec*) and dynamic contextual embeddings (from *BERT*). Acoustic and linguistic features are also included. We perform experiments on the publicly available dataset TED-LIUM mixed with real noise. The proposed rescoring approaches give significant improvement of the WER over the ASR system without rescoring models in two noisy conditions and with n-gram and RNNLM.

## 1. INTRODUCTION

ASR systems have made significant progress in recent years. However, the ASR WER increases then the system conditions, in which they are trained, differ from those in which they are used. Classical ASR systems take into account only acoustic (acoustic model), lexical, and syntactic information (local n-gram language models (LM)). In the case of mismatch conditions between training and testing, like noise, the signal is distorted and the acoustic model can be not able to compensate for this variability. Even if the noise compensation methods work well [9], it may be interesting to *incorporate semantic knowledge* in the decoding process to help the ASR to combat adverse conditions. Indeed, this information is important for ASR systems. Some studies have tried to include such information into an ASR. The authors of [14] use a semantic context for recovering proper names missed in the ASR process. [1] integrate semantic frames and target words into recurrent neural network LM. In [2], the re-ranking of the ASR hypotheses using an in-domain LM and a semantic parser can significantly improve the accuracy of transcription and semantic understanding. [5] introduce the semantic grammars applicable for ASR and understanding using ambiguous context information.

*Rescoring the ASR N-best hypotheses list* can be an efficient solution to incorporate long-range semantic information. [16] formalize the N-best list rescoring as a learning problem and use a wide range of features with automatically optimized weights to re-rank the N-best lists. [12] introduce N-best rescoring through long short-term memory (LSTM)-based encoder network followed by a fully-connected feedforward NN-based binary-class classifier. [15] propose a bidirectional LM for rescoring, and utilizes the word prediction capability of the *BERT* model [3][19] using the masked word tokens.

In this work, we aim to *add long-range semantic information to ASR through the rescoring the ASR N-best hypotheses list in the specific context of noisy data*. We believe that some ASR errors can be corrected by taking into account distant contextual dependencies. This is especially important for noisy conditions. We hope that in very noisy parts of speech, the semantic model could help to remove the acoustic ambiguities. The core ideas of the proposed rescoring approaches are:

(a) To rescore the ASR N-best list using two types of continuous semantic models to represent each hypothesis: static word-based *word2vec* [11] and dynamic sentence-based *BERT* [3][19]. Compared to [15], where only one sentence is taken at inference and so *masked word prediction* is performed for *BERT*, we use the *sentence prediction capability* of the *BERT* model and compare ASR hypotheses two per two. We compare the short-term *word2vec* model with the long-range *BERT* one.

(b) To represent each hypothesis by a semantic model and to compare two per two. For each pair of hypotheses, we provide two scores. Compared to the two-per-two comparison of [12], where all the comparisons are performed between the oracle hypothesis and one of the other selected hypotheses, we compare all hypotheses from the N-best list. This allows us to not miss a good hypothesis.

(c) To combine these scores with the ASR scores attached to each of the hypotheses (acoustic and linguistic)

and to use the combined scores to rescore the ASR N-best list hypotheses.

(d) We also propose efficient DNN-based procedures for training the proposed semantic hypotheses representations. Acoustic and linguistic scores are incorporated in this DNN.

Compared to [16] we use different DNN features. Compared to our previous work [8], we use the powerful *BERT*-based semantic model, represent the hypotheses at the sentence level, and train hypotheses representations by a DNN.

In experiments using a publicly available speech corpus, we systematically explore the effectiveness of proposed features and their combinations. The proposed approaches steadily outperform the classical ASR system.

## 2. PROPOSED METHODOLOGY

### 2.1. Introduction

A classical speech recognition system provides an acoustic score $P_{ac}(w)$ and a linguistic score $P_{lm}(w)$ for each of the hypothesized word $w$ of the sentence to recognize. The best sentence hypothesis is the one that maximizes the likelihood of the word sequence:

$$\widehat{W} = argmax_{h_i \in H} \prod_{w \in h_i} P_{ac}(w)^\alpha * P_{lm}(w)^\beta \qquad (1)$$

$\widehat{W}$ is the recognized sentence (the end result); $H$ is the set of N-best hypotheses; $h_i$ is the *i*-th sentence hypothesis; $w$ is a hypothesized word. α and β represent the weights of the acoustic and the LMs.

An efficient way to take into account semantic information is to re-evaluate (rescore) the best hypotheses of the ASR system. In [8] we proposed to introduce the semantic probability for each word $P_{sem}(w)$ to take into account the semantic context of the sentence. This was performed through a definition of *context part* and *possibility zones*. In this rescoring approach, $P_{ac}(w)$, $P_{lm}(w)$, and the semantic score $P_{sem}(w)$ are computed separately and combined using specific weights α, β and γ (for $P_{sem}(w)$) for each hypothesis.

In this work, we propose to better represent these scores and to go beyond a simple score combination. We propose a DNN-based rescoring model that rescores a pair of ASR hypotheses, one at a time. Each of these pairs is represented by a feature vector including acoustic, linguistic, and semantic information. This vector is used as DNN input for the rescoring model. We use hypotheses pairs to get a tractable size of the DNN input vectors. In our approach, semantic information is introduced using two types of representations: *word2vec* and *BERT*. Different semantic properties and efficiencies of these representations motivate us to explore them for our task of ASR in noisy conditions.

### 2.2. DNN-based rescoring model

The main idea behind our rescoring approach is (a) to train a DNN-based rescoring model with input features extracted from the ASR N-best list, (b) to rescore the N-best hypotheses, and (c) to select the hypothesis with the best score as the recognized sentence. We first present the rescoring scheme. The DNN input features will be presented in the following subsection.

As mentioned before, our DNN-based rescoring model rescores per pairs of ASR hypotheses. We decided to put at the input of the DNN the features computed from a pair of hypotheses. For each pair of hypotheses $(h_i, h_j)$, the expected DNN output is:
- *1,* if word error rate (WER) of $h_i$ is lower than WER of $h_j$;
- *0,* otherwise.

The overall algorithm of the N-best list rescoring is as follows. For a given sentence, for each hypothesis $h_i$ we want to compute the cumulated score *score($h_i$)*. To perform this, for each hypotheses pair $(h_i, h_j)$ of the N-best list of this sentence:
- we apply the DNN rescoring model and obtain the output value *v* (between 0 and 1). The value *v* close to 1 means that the $h_i$ is better than $h_j$. We use this value to compute the scores for these hypotheses.
- we update the scores of both hypotheses as:
  score($h_i$) += v;   score($h_j$) += 1-v.

At the end, the hypothesis that obtains the greatest score is chosen as the recognized sentence.

*2.2.1 word2vec-based rescoring method*

For this method, we define the *context part* and the *possibility zones* of the N-best list. *A context part* consists of the words which are common to all the N-best hypotheses generated by the ASR. We assume that this part captures the semantic information of the topic context of the document. We represent the context part with the average of the *word2vec* embedding vectors of the words of the context part [14], [7]:

$$V_{context} = \sum_{w \in context} V_{word2vec}(w) / nbrw_{context} \qquad (2)$$

where $nbrw_{context}$ is the number of words in the context part.

The *possibility zones* of a hypothesis are the set of words that do not belong to the context part. Possibility zones correspond to the area where we want to find the words to be corrected. We represent the possibility zones of each hypothesis by the average of the *word2vec* embedding vectors of the words of the possibility zones:

$$V_{hi} = (\sum_{\substack{w \in hi \\ w \notin context}} V_{word2vec}(w)) / nbrw_{poss} \qquad (3)$$

where $nbrw_{poss}$ is the number of words in the possibility zones.

For a pair of hypotheses $(h_i, h_j)$, the input vector of the proposed *word2vec*-based DNN rescoring model could contain the following features:
- context part vector $V_{context}$ ;

- possibility part vector $V_{hi}$ for hypothesis $h_i$;
- possibility part vector $V_{hj}$ for hypothesis $h_j$;
- cosine distance between $V_{context}$ and $V_{hi}$;
- cosine distance between $V_{context}$ and $V_{hj}$;
- acoustic score of $h_i$: $P_{ac}(h_i) = \prod_{w \in h_i} P_{ac}(w)$;
- acoustic score of $h_j$: $P_{ac}(h_j) = \prod_{w \in h_j} P_{ac}(w)$;
- linguistic score of $h_i$: $P_{lm}(h_i) = \prod_{w \in h_i} P_{lm}(w)$;
- linguistic score of $h_j$: $P_{lm}(h_j) = \prod_{w \in h_j} P_{lm}(w)$.

The advantage of this method is that we can model together semantic, acoustic, and linguistic scores of each hypothesis. In this case, we do not need explicit weighting coefficients α, β, and γ.

*2.2.2 BERT-based rescoring method*
*BERT* is a multi-layer bidirectional transformer encoder that achieves state-of-the-art performances for multiples natural language tasks. The pre-trained *BERT* model can be fine-tuned using task-specific data [17].

As the cosine distance is not meaningful for *BERT* semantic model [20][21], we cannot use it to compare the context part and possibility zones of the hypotheses, as we did with the *word2vec* model. So, we only compute the semantic information at the sentence level, as described below.

In our approach, we propose to use pre-trained *BERT* models and fine-tune them using our application-specific data. Two methods can be used to fine-tune *BERT*: *masked LM* and *next sentence prediction*. We based our *BERT* fine-tuning on a task similar to the latter one. We input a hypotheses pair $(h_i, h_j)$, that we want to compare, to a BERT model. The output is set to 1 (or 0) if the first (or the second) hypothesis achieved the lowest WER. During fine-tuning, the hypotheses pairs that get the same WER are not used. During evaluation (with development and test sets), all hypotheses pairs are considered.

## 3. EXPERIMENTAL CONDITIONS

### 3.1. Corpus description

We used the TED-LIUM corpus [4], containing the recordings of the TED conferences. This corpus is publicly available. Each conference is focused on a particular subject and so the corpus is well suited to our study. We used the train, development and test partitions provided with the TED-LIUM corpus: 452 hours for training (268k segments, 4778k words, 452h), 8 conferences (507 segments, 17783 words, 1h36) for development, and 11 conferences (1155 segments, 27500 words, 2h37) for test set (see Table 1).

This research work was carried out as part of an industrial project, concerning the recognition of speech in noisy conditions, more precisely in a fighter aircraft. So, we added noise to the development and test sets to get closer to real aircraft conditions: additive noise at 10 dB and 5dB SNR (noise of a F16 from the NOISEX-92 corpus [18]). The noise is not added to the training part.

| Data | Nbr. of talks | Nbr. of words | Duration |
|---|---|---|---|
| Train | 2351 | 4778k | 452h |
| Development | 8 | 17783 | 1h36 |
| Test | 11 | 27500 | 2h37 |

**Table 1**. The statistics of the TED-LIUM dataset.

### 3.2. Recognition system description

We used a recognition system based on the Kaldi voice recognition toolbox [13]. TDNN triphone acoustic models are trained on the training part (without noise) of TED-LIUM. The lexicon and LM were provided in the TED-LIUM distribution. The lexicon contains 150k words. The LM has 2 million 4-grams and was estimated from a textual corpus of 250 million words. We also performed recognition using the RNNLM model [10]. We want to see if using more powerful LM, the proposed rescoring models can improve the ASR. In all experiments, during rescoring, the LM (4-gram or RNNLM) is not modified.

As usual, we used the development set to choose the best parameter configuration and the test set to evaluate the proposed methods with this best configuration. We used the WER to measure the performance. The WER of our ASR system on TED-LIUM publicly provided test set is around 8 % using n-gram LM (using the training and the test sets without added noise).

According to our previous work on the semantic model [8], the use of 5 or 10 hypotheses of the N-best list is not enough for the efficient semantic model. Using more than 25 hypotheses shows no further improvement. In the current work, we chose to use the N-best list of 20 hypotheses in all our experiments. Moreover, this size of the N-best list seems to be reasonable to generate the pairs of hypotheses and to have the tractable computational load during the training of rescoring models.

### 3.3 *word2vec* embeddings

We trained *word2vec* model on a text corpus of a billion words extracted from the *OpenWebText* corpus. The size of the generated embedding vector is 300 and the embedding models 700k words. As DNN configuration to training the *word2vec*-based rescoring model, we used a neural network with 2 hidden layers. For the last layer, the activation function is a sigmoid and the loss function is a binary cross-entropy.

### 3.4 *BERT* models

We downloaded the pre-trained *BERT* models provided by Google [17]. We performed the experiments using models with 4, 8, or 12 layers and the size of the hidden layers is 128, 256, or 512 neurons. In the tables, we note these models as *LxxHyyy*. For instance, *L8H256* means the *BERT* model with 8 hidden layers and 256 as the size of each hidden layer.

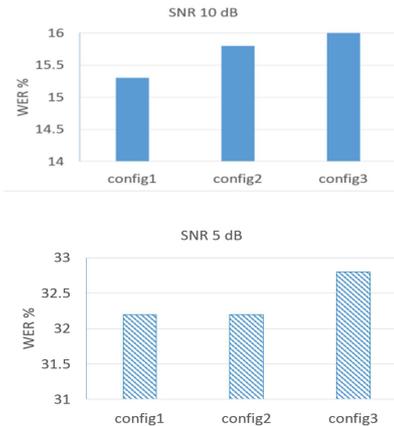

**Figure 1**. ASR WER (%) on the TED-LIUM development corpus for different *Word2vec* model configurations. The top bar chart corresponds to SNR of 10 dB, bottom to SNR of 5 dB. 4-gram LM. *word2vec-based* DNN model is trained using 1 M hypotheses pairs. Configurations represent different DNN input features.

## 4. EXPERIMENTAL RESULTS

### 4.1 Impact of hyperparameters

In this section, we investigate the different hyperparameters of the proposed *word2vec*-based and *BERT*-based rescoring models. These hyperparameters are studied on the development set of TED-LIUM and the best values were applied for the final evaluation on the test set. We use 4-gram LM for recognition. As mentioned previously, during rescoring the LM is not modified.

*4.1.1. Word2vec-based rescoring model*
**Impact of the size of the training corpus for DNN-based rescoring.** We used three different sizes of the training data: 1 million of pairs of hypotheses (corresponding to 100 TED-LIUM talks of the training set, without noise), 6.6 million of pairs of hypotheses (500 TED-LIUM talks of the training set) and 13.2 million of pairs of hypotheses (1000 TED-LIUM talks of the training set). We observed the same performance of the *word2vec*-based model for all sizes of data.

**Impact of different DNN input features.** We evaluate three configurations (*config1, config2, config3* in Figure 1): for configuration 1 (*config1*), the DNN input contains only acoustic scores differences, linguistic scores differences and cosine difference for each hypotheses pair (3 features); in configuration 2 (*config2*), we use the acoustic, linguistic scores and cosine distances (6 features); configuration 3

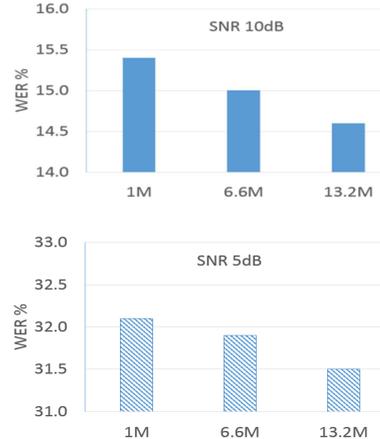

**Figure 2**. ASR WER (%) on the TED-LIUM development corpus as function of the amount of *BERT* fine-tuning data. The top bar chart corresponds to SNR of 10 dB, bottom to SNR of 5 dB. 4-gram LM, *L8H128* BERT model.

(*config3*) implements all input features, presented in Section 2.2.1 (906 features). According to Figure 1, configuration 1 achieves the best performance. Surprisingly, configuration 3 performs worse than configuration 1, indicating that embedding features provide no benefit. It is possible, that the relevant acoustic and linguistic data are diluted because the size of the embedding features (900) tends to dominate the size of the acoustic and linguistic features (4).

In the following experiments, a *word2vec*-based rescoring model based on only 1 million of hypotheses pairs for training and configuration 1 of the DNN-based input features will be used.

*4.1.2 BERT-based rescoring model*
In this section, we study the *BERT*-based rescoring model. Acoustic and LM probabilities are not used in these experiments. They will be used in the overall evaluation.

**Impact of the size of the training corpus for DNN-based rescoring.** Figure 2 presents the results on the development corpus using L8H128 *BERT* rescoring model with different amounts of data, i.e. pairs of N-best hypotheses, for fine-tuning. As acoustic and linguistic probabilities are not used here, fine-tuning data is text data. The figure in the top corresponds to the results for SNR 10 dB and in the bottom for SNR 5 dB. These results show that increasing the size of the fine-tuning data has a significant effect on the WER: more fine-tuning data is profitable to obtain an efficient *BERT*-based semantic model.

**Impact of the number of hidden layers of the BERT model.** Figure 3 shows the recognition performances as a function of the number of layers of the BERT model. The size of the hidden layers is 128 and the size of the fine-tuning data is 13.2M hypotheses pairs. Using 12 layers gives the best performance for the two SNR levels. We observe that this parameter plays an important role.

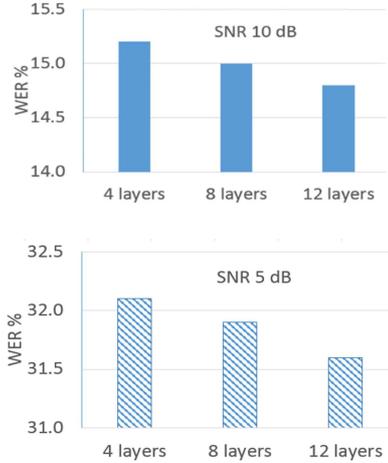
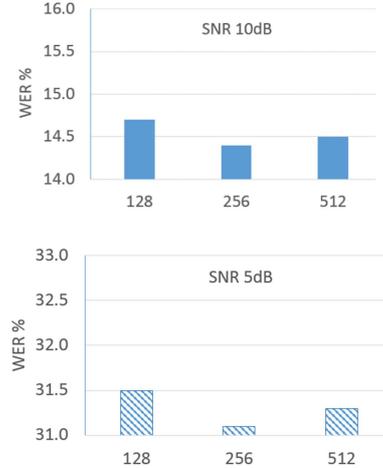

**Figure 3**. ASR WER (%) on the TED-LIUM development corpus according to the number of layers for the *BERT* model. The top bar chart corresponds to SNR of 10 dB, bottom to SNR of 5 dB. 4-gram LM, *LxH128 BERT* model fine-tuned using 13.2M hypotheses pairs.

**Figure 4**. ASR WER (%) on the TED-LIUM development corpus as function of the size of hidden layer of *BERT*. The top bar chart corresponds to SNR of 10 dB, bottom to SNR of 5 dB. 4-gram LM, *L12Hyyy BERT* semantic model fine-tuned on 13.2M hypotheses pairs.

*Impact of the size of the hidden layers.* Figure 4 reports the importance of the size of the hidden layers. We used the *L12Hyyy BERT* model fine-tuned on 13.2M hypotheses pairs. We can observe a variation according to the size of the hidden layers. The best performance is obtained for a size of 256 and will be used in the following.

In conclusion, we can say that for the *BERT*-based rescoring model, it is important to use a large corpus for fine-tuning the model and to choose a model with many hidden layers. The size of 256 for hidden layers seems to be a good compromise.

### 4.2 Overall results

To further analyze the impact of proposed rescoring models, Table 2 reports the WER for the development and the test sets of TED-LIUM with noise conditions of 10 and 5 dB. In the table, the first line of results (method *Random*), corresponds to the random selection of the recognition result from the N-best hypotheses. This is performed without the use of the proposed rescoring models. The second line of the table (method *Baseline*), corresponds to not using the rescoring models (standard ASR). The last line of the table (method *Oracle*) represents the maximum performance that can be obtained by searching in the N-best hypotheses: we select the hypothesis which minimizes the WER for each sentence. The other lines of the table give the performance of the proposed approaches. For all experiments, the N-best list of 20 is used.

For *BERT*-based rescoring model, we studied 3 configurations:

- rescoring using only the scores *score(h)* computed with *BERT*-based rescoring method as presented in the section 2.2.2 (denoted *BERT* in the tables);
- rescoring using a combination of the *BERT*-based score *score(h),* and the acoustic score $P_{ac}(h)$ (*BERT with ac. score* in the tables). In this case, *score(h)* is used as a *pseudo probability* and multiplied to the acoustic likelihood with a proper weighting factor (to optimize);
- rescoring using a combination of the *BERT*-based score, the acoustic score $P_{ac}(h)$ and the linguistic score $P_{lm}(h)$ (*BERT with ac./ling. scores* in the tables). This combination is performed as previously.

We present the results only for the best *BERT* model *L12H256* fine-tuned using 13.2 million pairs of hypotheses (see section 4.1).

From this table, we can make the following observations. For all conditions and all evaluated rescoring models, the proposed rescoring models outperform the baseline system. This shows that the proposed rescoring models are efficient to capture a significant proportion of the semantic information. For all datasets, *word2vec with ac./ling. scores* configuration gives a small but significant improvement compared to the baseline system.

Unsurprisingly, the proposed *BERT*-based rescoring models outperform the *word2vec*-based model. It is important to note that in *word2vec*, the word embedding is static and the words with multiple meanings are conflated into a single representation. In the BERT model, the word embedding is dynamic and so more powerful because one word can have several embeddings in the function of the context words. Adding the acoustic score to the *BERT* rescoring model (*BERT with ac. score* in the tables) improves the performance. Indeed, the acoustic score is an important feature and should be taken into account. On the other hand, adding the linguistic score alone to the *BERT* rescoring gives no improvement compared to the *BERT* model. We do not present this result in the tables. Using the linguistic and acoustic scores in the *BERT* rescoring model (*BERT with ac./ling. scores*) brings only small improvement compared to *BERT with ac. score*: Google's BERT model, trained on the billions of sentences, probably captures the linguistic structure of the language better than a simple n-gram LM trained on a much smaller corpus.

| SNR 10 dB | | |
|---|---|---|
| Method | Dev | Test |
| Random | 16.9 | 22.9 |
| Baseline system | 15.7 | 21.1 |
| *word2vec with ac./ling. Scores* | 15.3 | 20.6 |
| *BERT* | 14.4 | 19.8 |
| *BERT with ac. score* | 14.2 | **19.4** |
| *BERT with ac./ling. Scores* | **14.1** | **19.4** |
| Oracle | 11.2 | 15.0 |
| SNR 5 dB | | |
| Method | Dev | Test |
| Random | 33.5 | 41.3 |
| Baseline system | 32.7 | 40.3 |
| *word2vec with ac./ling. Scores* | 32.1 | 39.2 |
| *BERT* | 31.1 | 38.7 |
| *BERT with ac. score* | **30.6** | **37.9** |
| *BERT with ac./ling. Scores* | **30.6** | **37.9** |
| Oracle | 27.5 | 33.2 |

**Table 2**. ASR WER (%) on the TED-LIUM development and test sets, SNR of 10 and 5 dB. N-best hypotheses list of 20 hypotheses, 4-gram LM. *L12H256 BERT* model fine-tuned on 13.2M hypotheses pairs.

| SNR 10 dB | | |
|---|---|---|
| Method | Dev | Test |
| Random | 13.9 | 20.2 |
| Baseline system | 12.3 | 17.7 |
| *word2vec with ac./ling. scores* | 12.0 | 17.5 |
| *BERT* | 12.0 | 17.4 |
| *BERT with ac. score* | 11.6 | 17.1 |
| *BERT with ac./ling. scores* | **11.5** | **16.9** |
| Oracle | 8.3 | 12.1 |
| SNR 5 dB | | |
| Method | Dev | Test |
| Random | 29.2 | 38.4 |
| Baseline system | 28.2 | 37.1 |
| *word2vec with ac./ling. scores* | 27.4 | 36.3 |
| *BERT* | 27.0 | 35.9 |
| *BERT with ac. score* | 26.6 | **35.3** |
| *BERT with ac./ling. scores* | **26.5** | 35.4 |
| Oracle | 23.1 | 30.2 |

**Table 3**. ASR WER (%). N-best hypotheses list of 20 hypotheses. TED-LIUM development and test sets, SNR of 10 and 5 dB, RNNLM. *L12H256 BERT* model fine-tuned on 13.2M hypotheses pairs.

For *BERT*-based results, all improvements are significant compared to the baseline system (confidence interval is computed according to the matched-pairs test [6]). On the test set, *BERT with ac./ling. scores* obtains an absolute improvement of 1.7 % for 10 dB (19.4 % WER versus 21.1 % WER) and 2.4 % for 5 dB (37.9 % versus 40.3 %) compared to the baseline system. This corresponds to about 8 % (for 10 dB) and to about 6 % (for 5 dB) of relative improvement.

As n-gram LM is limited in its ability to model language context (long-range dependencies), we performed the ASR experiments using more powerful RNNLM. Table 3 reports the results for the same set of experiments, but, instead of n-gram, the RNNLM is used during the speech recognition. The proposed rescoring methods give consistent improvements as for n-gram LM results. So, all previous observations are valid for RNNLM-based experiments. The best system (*BERT with ac./ling. Scores*) gives about 4.5 % for 10 dB (16.9 % versus 17.7 %), and about 4.6 % for 5 dB (35.4 % versus 37.1 %) of relative improvement compared to the baseline system. These improvements are significant, except for the test set 10 dB. In the case of RNNLM, the improvement is smaller compared to the 4-gram case. It is possible that RNNLM reduces the effect of semantic rescoring.

## 5. CONCLUSION

In this article, we focus on the task of automatic speech recognition in noisy conditions. Our methodology is based on taking into account semantics through representations that capture the semantic characteristics of words and their context. The semantic information is taken into account through a rescoring module on ASR N-best hypotheses. We proposed two effective approaches: *word2vec*-based and *BERT*-based. The information extracted thanks to these representations is trained using DNN-based learning. Acoustic and linguistic information is integrated too. To evaluate our methodology, the corpus of TED-LIUM conferences corrupted with real noise is used. The best system *BERT with ac./ling. scores* gives about 8 % (4-gram) and 4.5 % (RNNLM) for 10 dB; about 6 % (4-gram) and 4.6 % (RNNLM) for 5 dB of relative improvement compared to the baseline system. These improvements are statistically significant, except for the test 10 dB RNNLM.

In future work, we would like to improve the *BERT*-based rescoring method by integrating acoustic and linguistic scores in the fine-tuning.

## 6. ACKNOWLEDGMENTS

The authors thank the DGA (*Direction Générale de l'Armement*, part of the *French Ministry of Défence*), Thales AVS and Dassault Aviation who are supporting the funding of this study and the "*Man-Machine Teaming*" scientific program in which this research project is taking place.